\pdfoutput=1
\documentclass[11pt]{article}

\usepackage[table]{xcolor}

\usepackage[final]{acl}

\usepackage{times}
\usepackage{latexsym}

\usepackage[T1]{fontenc}
\usepackage[utf8]{inputenc}

\usepackage{microtype}

\usepackage{inconsolata}

\usepackage{graphicx}

\usepackage{booktabs}

\usepackage{enumitem}

\definecolor{lightred}{rgb}{1,0.9,0.9}
\definecolor{lightyellow}{rgb}{1,1,0.8}
\definecolor{lightgreen}{rgb}{0.9,1,0.9}
\definecolor{lightblue}{rgb}{0.9,0.92,1}

\usepackage{multirow}

\usepackage{arydshln}
\usepackage{listings}
\newcommand{\narrowtextsf}[1]{\textls[-50]{\textsf{#1}}}
\newcommand{\sys}[1]{\narrowtextsf{#1}}

\newcommand{\affilsup}[1]{\rlap{\textsuperscript{\normalfont#1}}}

\title{Infherno: \\End-to-end Agent-based FHIR Resource Synthesis \\from Free-form Clinical Notes}

\author{
    Johann Frei\affilsup{1}
    \qquad
    Nils Feldhus\affilsup{2,3,4}
    \qquad 
    Lisa Raithel\affilsup{2,3,4}
    \\
    \textbf{Roland Roller}\affilsup{4}
    \qquad
    \textbf{Alexander Meyer}\affilsup{2,5}
    \qquad
    \textbf{Frank Kramer}\affilsup{1}
    \\
    \scalebox{0.85}{
    $^1$IT-Infrastructure for Translational Medical Research, University of Augsburg} \\
    \scalebox{0.85}{
    $^2$BIFOLD – Berlin Institute for the Foundations of Learning and Data \quad
    $^3$Technische Universit\"at Berlin} \\
    \scalebox{0.85}{
    $^4$German Research Center for Artificial Intelligence (DFKI), Berlin} \\
    \scalebox{0.85}{
    $^5$IKIM, Charité - Universit\"atsmedizin Berlin
    }
}

\begin{document}
\maketitle
\begin{abstract}
% FHIR
For clinical data integration and healthcare services, the HL7 FHIR standard has established itself as a desirable format for interoperability between complex health data.
% Text-to-FHIR
Previous attempts at automating the translation from free-form clinical notes into structured FHIR resources address narrowly defined tasks and rely on modular approaches or LLMs with instruction tuning and constrained decoding.
% Motivation -> Solution
As those solutions frequently suffer from limited generalizability and structural inconformity, we propose an end-to-end framework powered by LLM agents, code execution, and healthcare terminology database tools to address these issues.
% Benefits
Our solution, called \sys{Infherno}, is designed to adhere to the FHIR document schema and competes well with a human baseline in predicting FHIR resources from unstructured text. The implementation features a front end for custom and synthetic data and both local and proprietary models, supporting clinical data integration processes and interoperability across institutions. Gemini 2.5-Pro excels in our evaluation on synthetic and clinical datasets, yet ambiguity and feasibility of collecting ground-truth data remain open problems.
\end{abstract}

\section{Introduction}

% Generic LLM introduction
Large language models (LLMs) have demonstrated strong performance in clinical and biomedical domains, as they have been shown to encode domain-specific knowledge~\cite{singhal-2023-llms-encode-clinical-knowledge,moor-2023-foundation-models-generalist-medical-ai}.
%
% Motivating IE application use case for clinical texts
They are increasingly used to answer clinical questions by processing relevant documents at inference time \cite{zakka-2024-almanac,chen-2025-challenging-medical-questions,wang-2024-direct}. However, this retrieval-based approach incurs significant latency and computational cost, as documents must be reprocessed for every query. This limits usability for tasks such as retrospective analysis or study planning with multiple queries over the same data \cite{coromilas-2021-covid-19-associated-arrhythmias,leibig-2022-breast-cancer-screening}.

A more scalable solution is to extract structured representations from clinical text in advance. If the extracted structure preserves the relevant information, it can be queried and reused instantly across multiple applications. 
% Target audience:
This is particularly important for healthcare service providers and clinical data integration efforts \cite{leroux-2017-semantic-interoperability,hong-2019-scalable-fhir,pimenta-2023-interoperability-clinical-data}.
% Introducing FHIR (briefly)
Here, the FHIR\footnote{\url{https://fhir.org/}} standard provides a flexible and interoperable format for representing healthcare data and is increasingly adopted to support standardized access to complex medical information.

\begin{figure}[t]
    \centering
    \resizebox{\columnwidth}{!}{%
	    \includegraphics{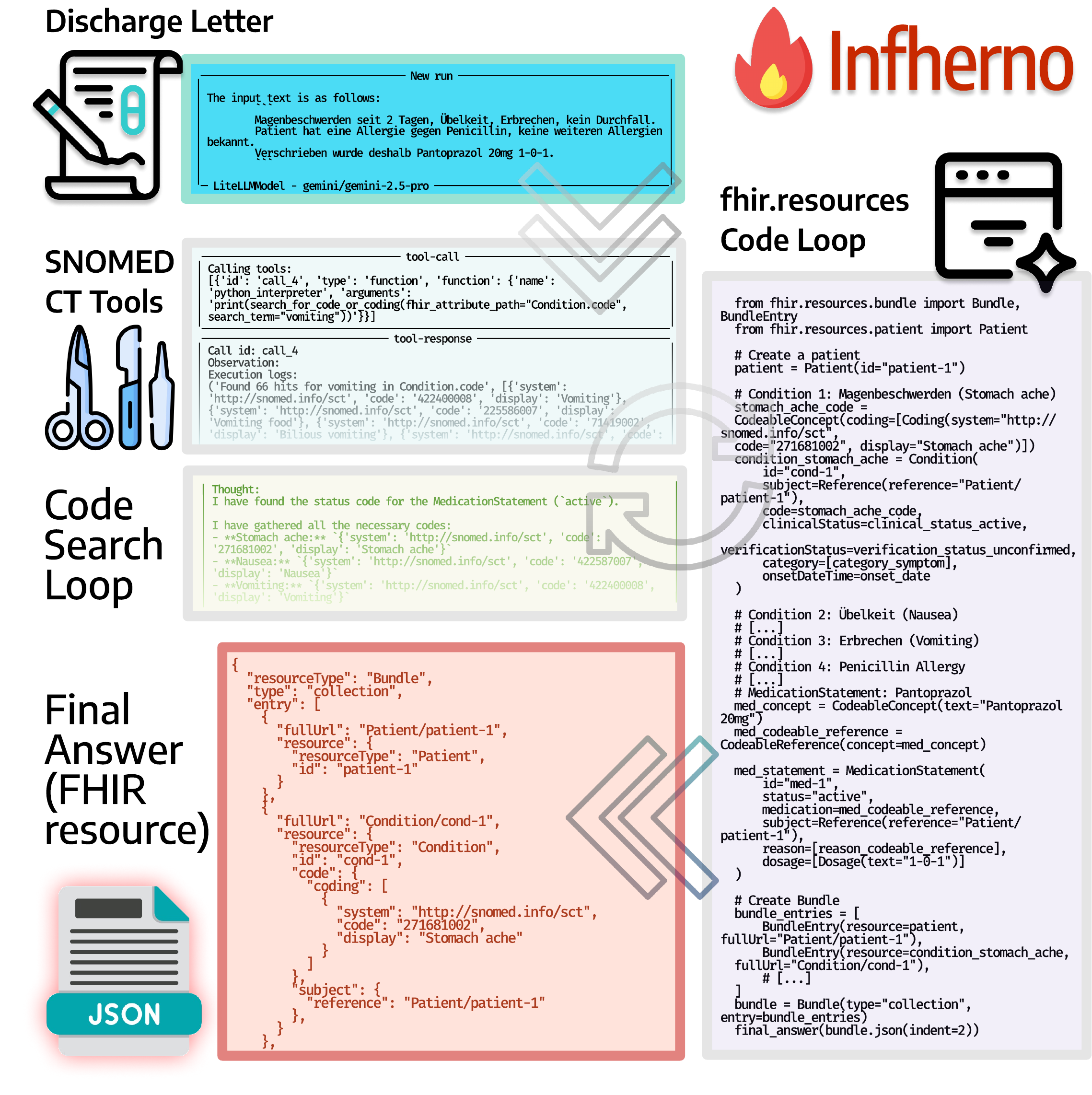}
    }
    \caption{Illustrative example of how \sys{Infherno}, an agentic approach for FHIR resource synthesis, processes a discharge letter (top left, cyan) using SNOMED CT tools (light blue) and terminology search (green) and \texttt{fhir.resources} code loops (purple, right). After a few iterations including tool calls and observations from a Python executor, the LLM agent proceeds to produce a final answer (red) in a FHIR/JSON format, representing the clinical information on patients and medications.}
    \label{fig:overview}
    %\vspace*{-1em}
\end{figure}

Conventional information extraction (IE) methods, such as classical named entity recognition (NER), relation extraction, and entity linking, are typically designed for narrowly defined tasks and fixed schemata. As such, they lack the flexibility to adapt to complex clinical contexts and often fail to produce complete, structured clinical representations.
In contrast, LLMs have shown promise for IE methods when they are framed as structured prediction tasks \cite{dagdelen-2024-structured-information-extraction}.
A recurring challenge in such tasks is ensuring that the output adheres to a specified schema, particularly when downstream components require well-structured inputs \cite{tavanaei-2024-so-lm}. This is especially true in domains like healthcare, where semantic correctness and schema compliance are essential.
Various approaches have been proposed to guide LLM outputs toward structural conformity, including fine-tuning, pre-training, instruction tuning, or constrained decoding \cite{shin-2021-constrained,geng-2023-gcd}. 
Some of them have dealt with text-to-FHIR translation \cite{sharma-2023-automated-digitization,li-2024-fhir-gpt,tabari-2025-enhancing-fhir,pope-patooghy-2025-fhir-proficiency}, but often encountered inconsistencies with the desired schema.
Agentic LLM approaches that ``reason'' through intermediate steps using external tools have emerged as a promising solution. Inspired by frameworks like Toolformer \cite{schick-2023-toolformer} and ReAct \cite{yao-2023-react}, models perform multiple tool-augmented reasoning steps, with validation and retry mechanisms to ensure correct output.

As our key contribution, we propose an end-to-end framework that transforms unstructured clinical text into rich, semantically accurate FHIR representations using an agentic LLM-approach. 
This enables holistic information extraction~\cite{zhang-2025-casereportbench, shao-2025-scalable-medication-extraction}, supports integration of both legacy and new data, and fosters interoperability across institutions.
Our contribution involves:
\begin{enumerate}[topsep=1pt, partopsep=0pt, itemsep=0pt, leftmargin=0.7cm]
\renewcommand{\labelenumi}{(\theenumi)}
    \item An end-to-end implementation for text-to-FHIR translation using LLM agents, SNOMED CT terminology integration and FHIR schema validation;
    \item Evaluation on real-life and synthetic data, with quantitative and qualitative error analyses across both proprietary and open-source LLMs to characterize failures and their severity;
    \item A lightweight demonstrator with front-end functionality, supporting both locally run and API-based state-of-the-art LLMs.
\end{enumerate}

\section{Background}

\paragraph{FHIR and SNOMED CT}

% What is FHIR
FHIR (Fast Healthcare Interoperability Resources) is a widely adopted standard for exchanging healthcare-related data, developed by the HL7 organization. FHIR defines resources as nested documents, often encoded in JSON, with well-defined types, required fields, enumerations, and references to other resources. A single FHIR resource can represent a broad range of entities, from patients and conditions to administrative structures like coverage or questionnaires.\footnote{See an example for a \texttt{Patient} resource object at: \url{https://hl7.org/fhir/R4/patient-example.json.html}}
% FHIR standard
FHIR facilitates the structured encoding of complex medical information in an interoperable fashion.

% FHIR CodeSystems & Bindings
A key feature of FHIR is the integration of internal and external code systems, composed as ValueSets to reference specific entities and concepts. Certain data elements may be constrained to a fixed, FHIR-internal code system to define the set of valid data values.\footnote{For instance, \texttt{Condition.clinicalStatus} only allows the values \texttt{active}, \texttt{recurrence}, \texttt{relapse}, \texttt{inactive}, \texttt{remission}, and \texttt{resolved}.} For certain fields, concepts can be referenced from external coding systems such as SNOMED CT\footnote{\url{https://www.snomed.org/what-is-snomed-ct}} or LOINC\footnote{\url{https://loinc.org/get-started/what-loinc-is/}}, and the set of valid data values can be further constrained by individual ValueSets, e.g., to limit data entries for body site to the subset of SNOMED CT concepts that only refer to body structures.
% FHIR Terminology Server
To search for codes and terms in a specific ValueSet, FHIR terminology servers provide a standardized interface for querying valid concepts. These servers commonly support multiple external code systems in addition to the FHIR-internal code systems.

% FHIR Ambiguity
While the FHIR schema is capable to accurately and verbosely capture complex clinical situations, it is also subject to structural and semantic ambiguity.
Practitioners often use only relevant subsets of data elements depending on their specific use cases. In addition, the standard does not always enforce the encoding of certain information into an unambiguous representation. For instance, a bone fracture of the left limb may be expressed as a \textit{Fracture of bone} SNOMED CT concept along with the \textit{bodySite} element referring to the \textit{Structure of left hand} concept, or purely by referring to the \textit{Fracture of bone of left hand} concept. Dosage information could be phrased only by a free-form text element, or by fully utilizing all relevant structured elements, rendering both approaches valid. Clinical notes may also be rather imprecise or ambiguous and require additional and subjective interpretation to fully infer the intended meaning, yet this issue also affects other, non-FHIR-based IE systems.
Therefore, comparing predicted and ground-truth FHIR data for semantic equivalence and correctness remains a non-trivial task.

\paragraph{Related Work}

\citet{sharma-2023-automated-digitization} presented a pipeline for digitizing prescription images into FHIR using separate components for extraction, normalization, and entity linking, limited to this particular task and mostly small-scale models.
\citet{li-2024-fhir-gpt} first applied LLMs to clinical text-to-FHIR transformation with human-annotated data\footnote{The human-annotated FHIR-GPT data has not been open-sourced to the best of our knowledge.}, but were limited to MedicationStatement resources and faced JSON parsing issues, whereas our validation loop ensures format conformity.
\citet{tabari-2025-enhancing-fhir} integrated a syntactic validator and zero- and few-shot strategies into their text-to-FHIR pipeline. Their setup is constrained to sentence-level conversion and exhibits less transparency due to the separation between the OpenAI model and the validator. In contrast, \sys{Infherno}'s tool-calling approach offers a higher degree of transparency and a larger variety of model choices.
\citet{pope-patooghy-2025-fhir-proficiency} explored a variety of FHIR-related tasks as a benchmark, but simplified them to short QA-style problems and also did not consider any elaborate pipeline with tools.
\citet{lee-2025-fhiragentbench} presented FHIR-AgentBench, a comprehensive benchmark for evaluating LLM agents on clinical question answering over FHIR-structured EHR data. Unlike \citeauthor{pope-patooghy-2025-fhir-proficiency}'s simplified tasks, they assess complex multi-step retrieval and reasoning over realistic FHIR resources, though their focus remains on querying existing data rather than generation.
\citet{idrissi-yaghir-2025-fhir-workbench} presented FHIR Workbench for evaluating text-to-FHIR generation, though models struggled without tool augmentation or validation mechanisms.
\citet{riquelme-tornel-2025-llms-for-automating-standardization} used GPT-4o and Llama-3.2 alongside clustering and retrieval generation approaches to perform automated FHIR mappings on MIMIC-IV (instead of free text), but missed out on evaluating the results manually.
Finally, \citet{schmiedmayer-2024-llm-on-fhir} aimed for an inverse perspective on the translation task by developing a mobile application that allows users to interact with FHIR resources via an LLM, while \citet{ehtesham-2025-mcp-fhir} presented an MCP-based agent for summarization and interpretation. Both represent a FHIR-to-text scenario which is focused on patient understanding.

\section{Infherno, an Agentic Approach}
Building on recent work on LLM agents in the medical domain \cite{liao-2025-reflectool, rose-2025-meddxagent, chen-2025-medbrowsecomp, wang-2025-baymax}, we propose an agentic framework that incorporates tool calls and coding to generate structured FHIR output from unstructured clinical text. 

The core task is to transform an unstructured clinical text into semantically corresponding FHIR representations.
Our approach follows the Thought-Code-Observation structure proposed as the ReAct framework by \citet{yao-2023-react}, and is implemented using the Smolagents~\cite{smolagents-2025} library which supports multi-step LLM agents with Python-code execution.
Figure~\ref{fig:overview} presents a simplified example of the \sys{Infherno} pipeline\footnote{Figure~\ref{fig:extended} in Appendix~\ref{sec:appendix} shows the extended version.}: Given a discharge letter, \sys{Infherno} which is equipped with tools accessing SNOMED CT, performing Terminiology Search, and executing Python code, is tasked to extract information pertinent to patients and medications. In the following, we describe each component:

\begin{figure}[t]
    \centering
    \resizebox{\columnwidth}{!}{%
	    \includegraphics[trim={2.25cm 3.25cm 2.25cm 0.25cm}, clip]{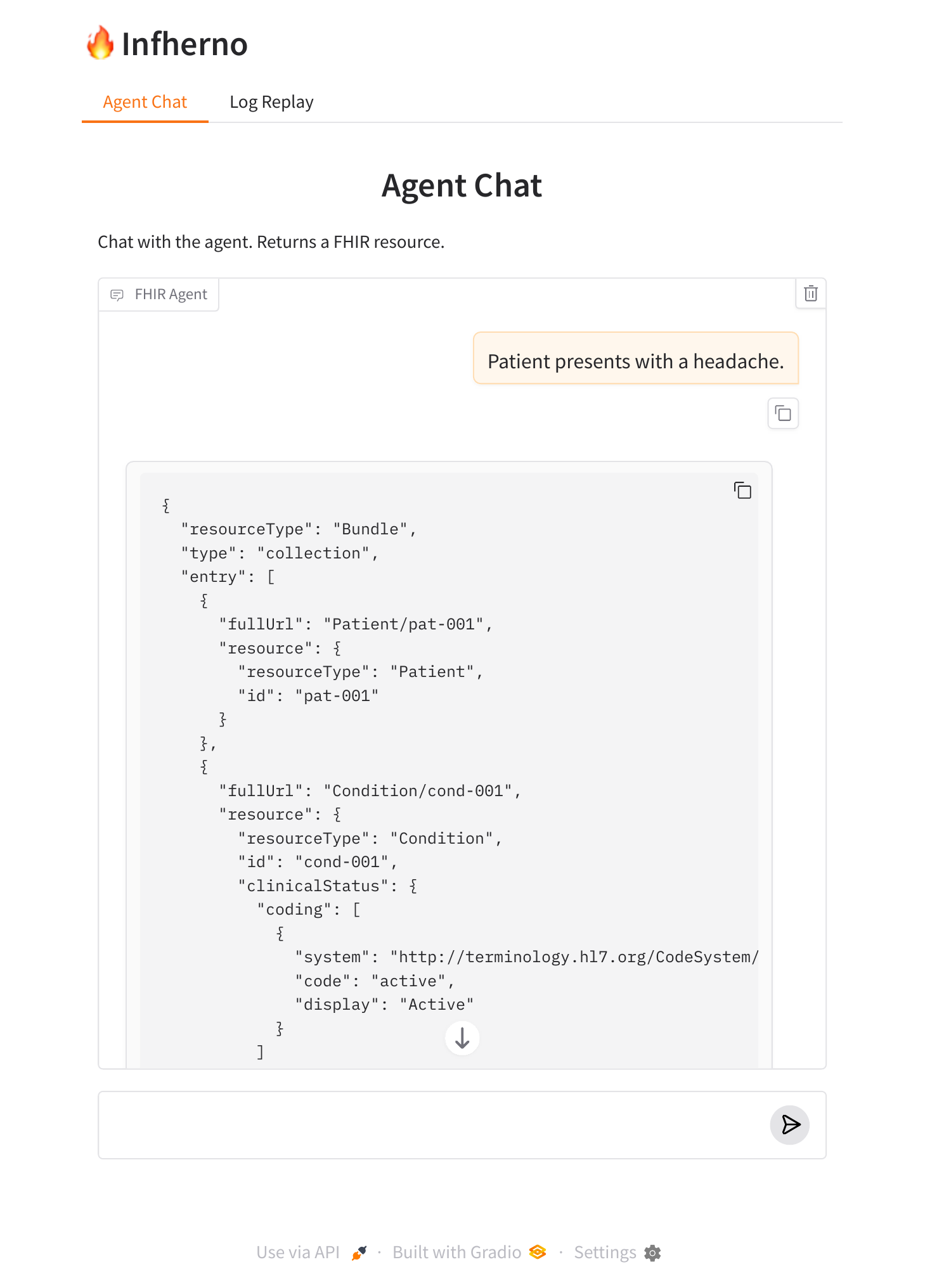}
    }
    \caption{Front end of \sys{Infherno} showing a short input text and the final answer as given by Gemini-2.5-Pro in the \textit{Agent Chat} function.}
    \label{fig:frontend}
    %\vspace*{-1em}
\end{figure}

\begin{figure}[t]
    \centering
    \resizebox{\columnwidth}{!}{%
	    \includegraphics[trim={0cm 2.75cm 0cm 0cm}, clip]{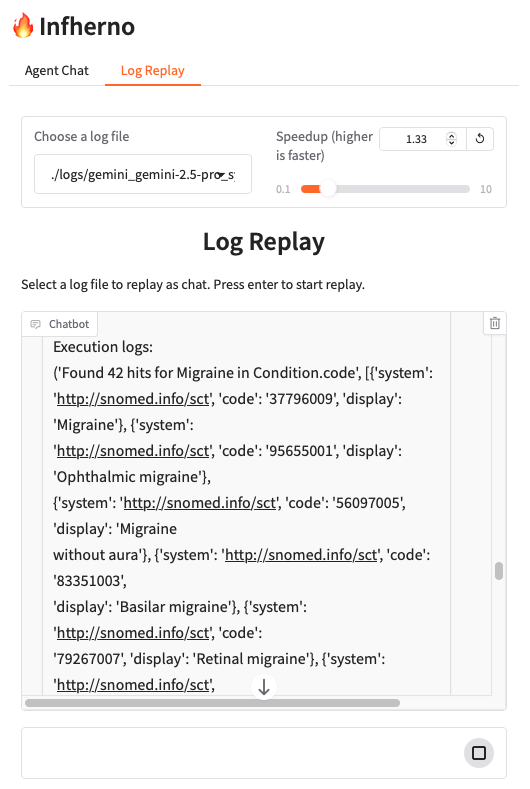}
    }
    \caption{Front end of \sys{Infherno} showing an intermediate step (Terminology Search) during the text-to-FHIR translation with the \textit{Log Replay} function.}
    \label{fig:logreplay}
    %\vspace*{-1em}
\end{figure}

\paragraph{Prompt Structure} To guide the agent's behavior, we include relevant contextual information into the prompt (Figure~\ref{fig:extended}, top left). This includes the unstructured input text, a list of target FHIR resource types, supported ValueSets, example code snippets demonstrating FHIR object creation, and a set of instructions on desired behaviors and constraints.

\paragraph{Terminology Search} To integrate FHIR-specific codes that conform to its specification, we provide our agentic system with an external, retrieval-augmented generation-based function to query particular terms in a set of supported FHIR ValueSets. This enables the agent to rely on external code systems, in particular the SNOMED CT ontology, to retrieve potential search results and include them into its context window. The external function call binds to an external FHIR terminology server to obtain a valid query response.

\paragraph{Structured Data as Code} Within the agent code execution stage, the agent is incentivized to use the \texttt{fhir.resources}\footnote{\url{https://github.com/nazrulworld/fhir.resources/tree/8.0.0}} Python module to create FHIR-conform data instances in a object-oriented fashion. This approach is crucial as it is able to catch morphological and syntactic errors early within the life cycle of the agent loop, and avoids cumbersome data validation that may arise from a purely JSON-centric FHIR document generation by the LLM. Since the library can directly provide error feedback, it can also facilitate the recovery from erroneous code predicted initially by the agent.

\paragraph{Output Formatting} As part of the Smolagents framework, the code agent can stop the agent loop by the \texttt{final\_answer} function call. Hereby, the agent is instructed to use the JSON-based object serialization of the \texttt{fhir.resources} module. This ensures that the response provides a structurally valid, FHIR-compliant JSON output. To deliver all generated FHIR resources to the user, the agent is instructed to aggregate them into a FHIR \textit{Bundle} that encapsulates the complete set.

\paragraph{Front End}
The visual interface of \sys{Infherno} is built on top of Gradio\footnote{\url{https://www.gradio.app/}} and allows the user to enter arbitrary clinical notes or select pre-defined examples from our synthetic dataset (Figure~\ref{fig:frontend}).
Intermediate steps including the tool calls and tool responses as well as the reasoning processes (``thoughts'' and ``observations'') of the ReAct framework \cite{yao-2023-react} are shown at inference time. A \textit{Log Replay} tab (Figure~\ref{fig:logreplay}) also enables to simulate the execution of already conducted experiments at custom speed without the need of an API key.
It supports the default Gemini API and OpenAI API via LiteLLM\footnote{\url{https://docs.litellm.ai/docs/}} and both API and local Hugging Face models -- a feature inherited from Smolagents.
The front-end app is available on \url{https://github.com/j-frei/Infherno}.

\section{Experiments}

To validate our agentic approach, we apply the agent to a set of medical documents to transform the unstructured text into a set of FHIR resources. Such individual comparison is highly complex due to the depth and richness of the FHIR schema and the complexity of clinical language.
Therefore, we conduct two experiments, involving both an automatic and a manual inspection and interpretation to assess the prediction quality.

\paragraph{Data}
For the experiments, we use two data sources.
We source ten anonymized English EHR documents from the n2c2 2018 challenge~\cite{henry_2018_2020} with drug-related NER annotations. The documents are used with focus on \texttt{MedicationStatement}, given the prior ground truth data. To craft ground-truth FHIR data, we first remove any drug-related NER spans if they do not refer to a medication instruction (e.g., drug allergies are not considered as such), normalize the drug-related NER span annotation into their common concepts to remove redundant mentions and create their corresponding \texttt{MedicationStatement} objects as ground truth.
As the n2c2 texts are anonymized, common Protected Health Information (PHI) elements like patient names do not occur and cannot be extracted. 

We also rely on ten ChatGPT-synthesized documents resembling medical discharge letters in German which include name and birth date mentions as well as described conditions and medication instructions, allowing us to share PHI-containing documents with remote LLM endpoints. The raw texts required manual editing. Obvious placeholder names, such as ``Max Mustermann'', were replaced in the synthetic texts to ensure realistic, non-repetitive patient names and addresses to improve the authenticity of the data.\footnote{The documents are publicly available on GitHub at \url{https://github.com/j-frei/Infherno}. One sample is included in the Appendix Figure~\ref{app:example_doc}.} To obtain suitable reference FHIR data, we annotate the documents from the corpus by manually extracting the relevant corresponding FHIR resources, referred to as human baseline (HB).
For these documents, we support \texttt{Patient}, \texttt{Condition}, and \texttt{MedicationStatement} as FHIR resources, since we consider these resource types to fit best to key clinical entities.

In general, the FHIR R4 release is targeted as it currently is the latest \textit{normative} release version.

\paragraph{Experimental Setup}
The first experiment aims to implement an evaluation requiring only minimal manual intervention or semantic interpretation. To measure the overlap between the predicted and ground-truth MedicationStatement concepts, we list and compare the \texttt{.medication} fields and manually categorize predicted concepts into TP, FP, FN to eventually calculate precision, recall and $F_1$ score. As the ground truth involves subjective interpretation of what drug mention is an actual medication instruction, we also add an \textit{relaxed} score setting that ignores certain ground truth entries that were jointly ignored or skipped by all LLMs. We also track the use of SNOMED CT codes and their correctness, as certain predictions only encode the concepts as text field. For the first experiment, both the n2c2 and synthetic data is used.

For the second experiment, we extensively compare the manual annotation and the generated annotation by verifying individual \textit{items} of each FHIR object across multiple FHIR resources. We define an item as a single unit of information, that may refer to, for instance, a single \texttt{birthDate} field but could also refer to a nested object item that describes a reference to a concept from an external coding system. Since the internal structure of certain objects is only meaningful in its entirety, we consider them as monolithic items in the evaluation, rather than decomposing their components. For this experiment, we only use synthetic data as it also covers synthetic PHI elements. Items are stratified into primary (e.g., \texttt{Condition.code}) and secondary (e.g., \texttt{Condition.verificationStatus}) items to account for the information importance differences. Evaluation decision details are highlighted in Appendix Section~\ref{app:second_experiment}.

\paragraph{Models}
For our agentic approach, we use both commercial and open-weight models for the first setup. This includes Claude Sonnet 4.5, Gemini-2.5 Pro\footnote{\url{https://ai.google.dev/gemini-api/docs/models}}, GPT-5 as well as DeepSeek V3.1, Qwen3-235B-A22B-2507, and Qwen3-8B.
As the second experiment incurs substantial manual effort, we select Gemini 2.5 Pro as target model as it performed best in the first setup on n2c2 data.

\section{Results \& Discussion}

\begin{table*}[t]
    \centering
    \resizebox{\textwidth}{!}{%
    \begin{tabular}{
      c |
      c |
      rrr
      rrr |
      rrr |
      c
    }
    \multicolumn{1}{c|}{\multirow{2}{*}{Data}}
    & \multicolumn{1}{c|}{\multirow{2}{*}{Model}}
        & \multicolumn{6}{c|}{Scores}
        & \multicolumn{3}{c|}{Eased Scores}
        & \multicolumn{1}{c}{Concepts}
    \\
    & 
        & TP & FP & FN
        & Pr & Re & $F_1$
        & FN & Re & $F_1$ 
        & total/with codes/with correct codes \\
    \toprule
    %\parbox[t]{2mm}{\multirow{6}{*}{\rotatebox[origin=c]{90}{\textbf{n2c2}}}}
    \multirow{6}{*}{\rotatebox[origin=c]{90}{\parbox{2cm}{\centering\textbf{n2c2}\\{\small27436 tokens}}}}
    & Claude Sonnet 4.5
        & 78 & 0 & 91
        & \textbf{1.0} & 0.462 & 0.632
        & 31 & 0.716 & 0.834
        & 79 / 78 / 78
        \\
    & DeepSeek V3.1 Chat
        & 82 & 0 & 85
        & \textbf{1.0} & 0.491 & 0.659
        & 25 & 0.766 & 0.868
        & 84 / 84 / 82
        \\
    & Gemini-2.5 Pro
        & 101 & 0 & 66
        & \textbf{1.0} & \textbf{0.605} & \textbf{0.754}
        & 6 & \textbf{0.944} & \textbf{0.971}
        & 104 / 104 / 104
        \\
    & GPT-5
        & 68 & 0 & 100
        & \textbf{1.0} & 0.405 & 0.576
        & 40 & 0.630 & 0.773
        & 69 / 63 / 42
        \\
    & Qwen3-235B-A22B-2507
        & 76 & 0 & 92
        & \textbf{1.0} & 0.452 & 0.623
        & 32 & 0.704 & 0.826
        & 77 / 76 / 76
        \\
    & Qwen3-8B
        & 32 & 0 & 134
        & \textbf{1.0} & 0.193 & 0.323
        & 74 & 0.302 & 0.464
        & 36 / 36 / 35
        \\
    \midrule
    %\parbox[t]{2mm}{\multirow{6}{*}{\rotatebox[origin=c]{90}{\textbf{synthetic}}}}
    \multirow{6}{*}{\rotatebox[origin=c]{90}{\parbox{2cm}{\centering\textbf{synthetic}\\{\small4065 tokens}}}}
    & Claude Sonnet 4.5
        & 13 & 1 & 3
        & 0.929 & 0.813 & 0.867
        & 2 & 0.867 & 0.897
        & 14 / 14 / 14
        \\
    & DeepSeek V3.1 Chat
        & 15 & 1 & 1
        & \textbf{0.938} & \textbf{0.938} & \textbf{0.938}
        & 0 & \textbf{1.0} & \textbf{0.968}
        & 16 / 16 / 16
        \\
    & Gemini-2.5 Pro
        & 14 & 1 & 1
        & 0.933 & 0.933 & 0.933
        & 0 & \textbf{1.0} & 0.966
        & 15 / 15 / 15
        \\
    & GPT-5
        & 15 & 1 & 1
        & \textbf{0.938} & \textbf{0.938} & \textbf{0.938}
        & 0 & \textbf{1.0} & \textbf{0.968}
        & 16 / 12 / 9
        \\
    & Qwen3-235B-A22B-2507
        & 15 & 1 & 1
        & \textbf{0.938} & \textbf{0.938} & \textbf{0.938}
        & 0 & \textbf{1.0} & \textbf{0.968}
        & 16 / 15 / 15
        \\
    & Qwen3-8B
        & 15 & 1 & 1
        & \textbf{0.938} & \textbf{0.938} & \textbf{0.938}
        & 0 & \textbf{1.0} & \textbf{0.968}
        & 16 / 15 / 15
        \\
    \bottomrule
    \end{tabular}
    }

    \caption{Evaluation scores from the first experiment, including Precision (Pr), Recall (Re), $F_1$ scores of \sys{Infherno} with various LLMs evaluated on n2c2 and synthetic data. Best scores are in \textbf{bold}.
    }
    \label{tab:scores}
    %\vspace*{-1.5em}
    
\end{table*}

\begin{table}[t]
    \centering
    \resizebox{\columnwidth}{!}{%
    \begin{tabular}{l|>{\columncolor{lightred}}r>{\columncolor{lightred}}r>{\columncolor{lightyellow}}r>{\columncolor{lightyellow}}r>{\columncolor{lightgreen}}r>{\columncolor{lightgreen}}r}
   \textbf{Category} & \multicolumn{2}{c}{\textbf{Worse than HB}} & \multicolumn{2}{c}{\textbf{Neutral}} & \multicolumn{2}{c}{\textbf{Better than HB}} \\
    \toprule
    
    importance & \texttt{prim} & \texttt{sec} & \texttt{prim} & \texttt{sec} & \texttt{prim} & \texttt{sec} \\
    \hdashline
    semantically related              &   0 &   4 &   0 &   4 &   0 &   0 \\
    completely identical              &   0 &   0 & 121 &  83 &   0 &   0 \\
    lacking in HB                     &   0 &  10 &   0 &  23 &  13 &  67 \\
    lacking in PD                     &   6 &  15 &   0 &  67 &   0 &   0 \\
    value difference                  &   0 &  10 &   0 &  12 &   5 &   1 \\
    semantic halluc. / invalid &   1 &   9 &   0 &   0 &   0 &   0 \\
    \midrule
    total               &  \multicolumn{2}{c}{\cellcolor{lightred} 46} & \multicolumn{2}{c}{\cellcolor{lightyellow} 314} &  \multicolumn{2}{c}{\cellcolor{lightgreen} 86} \\
    
    \bottomrule
    \end{tabular}
    }
    \vspace*{-0.5em}
    \caption{Second experiment: Results from the manual analysis between predicted (PD) and human baseline (HB) indicating the success and failure cases of \sys{Infherno} for \textbf{prim}ary and \textbf{sec}ondary items. Examples for different categories are shown in Appendix Table~\ref{app:tab:behavior_examples}.}
    \label{tab:scores_second_experiment}
    %\vspace*{-1em}    
\end{table}

According to Table \ref{tab:scores}, all LLMs demonstrate high precision with minimal fabrication\footnote{We found the FP in the synthetic data to be an error in the ground truth.}, but recall varies substantially on n2c2 data, where Gemini 2.5 outperforms all other models. The eased scores reveal that most failures represent cases systematically difficult for all models. For instance, drug information mentioned prior to or outside of the \texttt{DISCHARGE MEDICATION} section is often skipped, as it remains unclear whether such prior mentions should be considered superseded. Performance on synthetic data confirms that all models extract reliably when ambiguity is low.

Table \ref{tab:scores_second_experiment} presents detailed manual analysis comparing predictions against human annotations. Agreement is highest for primary items carrying essential clinical information. Notably, in nearly twice as many cases of quality differences, the model performed better than the human baseline rather than worse. Confabulations remain rare at the semantic level. Divergence occurs primarily on secondary items, which typically involve vague descriptions or ambiguous phrasing rather than explicitly stated primary elements.

\paragraph{Key Findings}
The validation highlights several important observations. First, the phrasing of the input text plays a critical role in annotation consistency. Vague or ambiguous expressions frequently lead to disagreements between the predicted and reference annotations, particularly for secondary items. In contrast, plainly stated and well-structured information is more reliably and consistently captured.

% Subjective Use of FHIR
Second, many divergences can be attributed to the partially subjective nature of FHIR in fringe cases. Minor or nonspecific health issues often fall into a gray area. These may either be excluded or encoded in different ways, such as as a \texttt{Condition} or an \texttt{Observation}. Since the experimental setup allowed only the use of \texttt{Patient}, \texttt{MedicationStatement}, and \texttt{Condition} resource types, the agent was not permitted to use the \texttt{Observation} resource, which limited some of its encoding options.

Furthermore, the \sys{Infherno} agent appears to be more cautious when deciding whether to encode uncertain symptoms. At the same time, it demonstrates stronger recall for clearly stated information that human annotators sometimes overlook, especially with a state-of-the-art LLM like Gemini 2.5. For example, the agent successfully included an address field that was missing in the human annotation. It also inferred an \texttt{onsetDateTime} by subtracting six weeks from the encounter date, which is a detail the human annotator did not encode.

These findings indicate that while human annotations are prone to fatigue and inconsistency, especially in repetitive and detail-oriented tasks, automated agents benefit from their ability to process dense text data using their large context as receptive field. As a result, they can achieve more reliable and comprehensive structured data extraction from our clinical text samples.

\section{Conclusion}
In conclusion, \sys{Infherno} presents a robust and effective framework and interface for transforming unstructured clinical data into standardized FHIR resources. Its agentic design, integrating external knowledge and validation, addresses critical challenges in clinical information extraction, paving the way for improved data interoperability in healthcare.
Future work includes the fine-tuning of smaller language models on the text-to-FHIR task and the integration of more FHIR resource types while further strengthening the robustness, and evaluate the approach on more diverse datasets.

\section*{Limitations}
\paragraph{Dataset and Annotations}
Manual evaluation of a larger, more diverse dataset was infeasible given the labor intensity and expertise required.
We rely on a single annotator for the human baseline, which inherently introduces a degree of subjectivity.

\paragraph{Resource Types}
Furthermore, our scope was intentionally limited to a subset of FHIR resource types (\texttt{Patient}, \texttt{Condition}, \texttt{MedicationStatement}).
Expanding to a broader range of FHIR resources would likely necessitate more verbose guidance in the system prompt, potentially increasing computational cost and latency.

\paragraph{Legal Remark}
Finally, from a legal perspective, it is important to note that Infherno interacts with a FHIR terminology server that includes an initialized SNOMED CT ontology. Therefore, a SNOMED CT license may be required if self-hosting a FHIR terminology server is desired.

\paragraph{CFG Baseline}
While we experimented with context-free grammar-based (CFG) approaches, we found that there are several reasons to object to this design choice:
\begin{itemize}[noitemsep,topsep=0pt,leftmargin=*]
    \item Creating a fully conformant FHIR grammar is a major engineering challenge.
    \item Applying a schema-based decoding may lead to an constrained decoding misalignment issue, which may result in divergences between constrained and unconstrained distributions, and may lead to generation instabilities and poor semantic outputs.
    \item While considering schema-based generation as an alternative to (\texttt{fhir.resources}-based) code-based FHIR Bundle assembly, other tasks like SNOMED CT code search must also be integrated into the pipeline process. There is no clear way of integrating all components in a non-agentic way. Orchestrating the pipeline in a non-agentic, multi-step pipeline flow rely on an inflexible, rigid process, which could negatively affect the final quality especially in complex clinical situations that do not fit well into a rigid data transformation process.
    \item An agentic-based flow allows for certain semantic cross-checks within a FHIR Bundle, that cannot be verified through a CFG-based approach. 
    \item In general, an agentic-based flow simplifies the addition of more (custom/user-defined) tools and validation checks.
\end{itemize}

\paragraph{Multilingual Support}
While one part of our evaluation focuses on a proprietary model (Gemini 2.5 Pro), our key aim of our work is to remain agnostic of specific LLMs. Consequently, support of other languages depends on the ability of the used LLM to process individual languages correctly rather than our system implementation. The system prompt is written in English and code switching is not used apart from the language in the input text. Since SNOMED CT is mostly an English system, English must be used to query for SNOMED CT codes.

\section*{Ethics Statement}
Depending on the selection of the LLM, we want to emphasize that users should be careful in selecting what data they enter. Most of the real-world medical datasets have licences and usage restrictions, so we recommend to use synthetic data only. Users should acknowledge the risk of leaking private data and de-identification.

\section*{Acknowledgments}
We thank the reviewers of the EACL 2026 and EMNLP 2025 System Demonstrations tracks for their valuable feedback. 
This research is funded by the Berlin Institute for the Foundations of Learning and Data (BIFOLD, ref. 01IS18037A) and the German Federal Ministry of Research, Technology and Space (MoMoTuBo, ref. FKZ01ZZ2008).

\bibliography{custom}

\appendix

\begin{table*}[t]
    \centering
    \resizebox{\textwidth}{!}{%
    	\begin{tabular}{rrrrr}
        \textbf{Level of equivalence} &  & & \textbf{\sys{Infherno}} (w/ Gemini-2.5-Pro) & \textbf{Human Baseline} \\
        \toprule
        
        \rowcolor{lightred} 
        \textbf{Worse than HB} 
            & 
            & 
            & 
            & \texttt{"code": "52795006",}
            \\
        \rowcolor{lightred} 
        (field not referenced) 
            & \texttt{[|+-?]}
            & Condition.bodySite
            & \texttt{"text": "Stirnbereich"}
            & \texttt{"display": "Forehead structure"}
            \\

        \rowcolor{lightred} 
        (hallucination)
            & \texttt{[X+?]}
            & Condition.category
            & \texttt{"code": "symptom"}
            & N/A
            \\

        \midrule
        \rowcolor{lightyellow}
        \textbf{Neutral} 
            & 
            & MS.dosage
            & N/A
            & \texttt{"code": "ordered",} \\
        \rowcolor{lightyellow}
        (optional field missing) 
            & \texttt{[-?]}
            & .doseAndRate.type
            & N/A
            & \texttt{"display": "Ordered"} \\

        \rowcolor{lightyellow}
        (total equivalence)
            & \texttt{[==]}
            & Condition.severity
            & \texttt{"code": "255604002",}
            & \texttt{"code": "255604002",}
            \\
        \rowcolor{lightyellow}
            & 
            & 
            & \texttt{"display": "Mild"}
            & \texttt{"display": "Mild"} 
            \\

        \midrule
        \rowcolor{lightgreen}
        \textbf{Better than HB} 
            & \texttt{[/+-!]}
            & Condition.code
            & \texttt{"code": "422400008",}
            & \texttt{"code": "422587007",} 
            \\
        \rowcolor{lightgreen}
        (inaccurate reference for \textit{throw up}) 
            & 
            & 
            & \texttt{"display": "Vomiting"}
            & \texttt{"display": "Nausea (finding)"}
            \\
        
    	\end{tabular}
    }
    \caption{Examples for level of equivalence and the manual validation between system output and human baseline.}
    \label{app:tab:behavior_examples}
    %\vspace*{-1em}
\end{table*}

\section{Quantitative Analysis}
\label{app:second_experiment}
We manually compared the items of each FHIR object in the human baseline annotation with their corresponding agent-generated equivalents. Missing items were tracked using \texttt{+} for those absent in the human baseline and \texttt{-} for those missing in the prediction. Potentially equivalent values were categorized as exact matches (\texttt{==}), semantically equivalent (\texttt{=}), or different (\texttt{+-}).

Items were tagged with \texttt{/} if the prediction value was preferable to the baseline, and with \texttt{|} if the baseline value was preferable. Items were left untagged when no clear preference could be determined or justified.

To distinguish essential from less relevant features during evaluation, we annotated core FHIR objects of major importance, such as patient information or the main diagnosis extracted from the input text, with \texttt{!} as primary items, and less critical elements, like vaguely described symptoms, with \texttt{?} as secondary items.
The importance level of a FHIR object determined the default importance of its internal items, unless overridden by manually applied, item-specific tags. These overrides were primarily used to demote non-essential items such as \texttt{Condition.verificationStatus} or \texttt{Patient.name.use}. Conversely, items from otherwise crucial FHIR resources, such as \texttt{Condition.subject} or \texttt{Condition.code}, were generally considered primary, as they carry core informational content.
Some examples are shown in Table~\ref{app:tab:behavior_examples}.

\section{Examples}
\label{sec:appendix}
\begin{figure*}[t]
    \centering
    \resizebox{\textwidth}{!}{%
	    \includegraphics{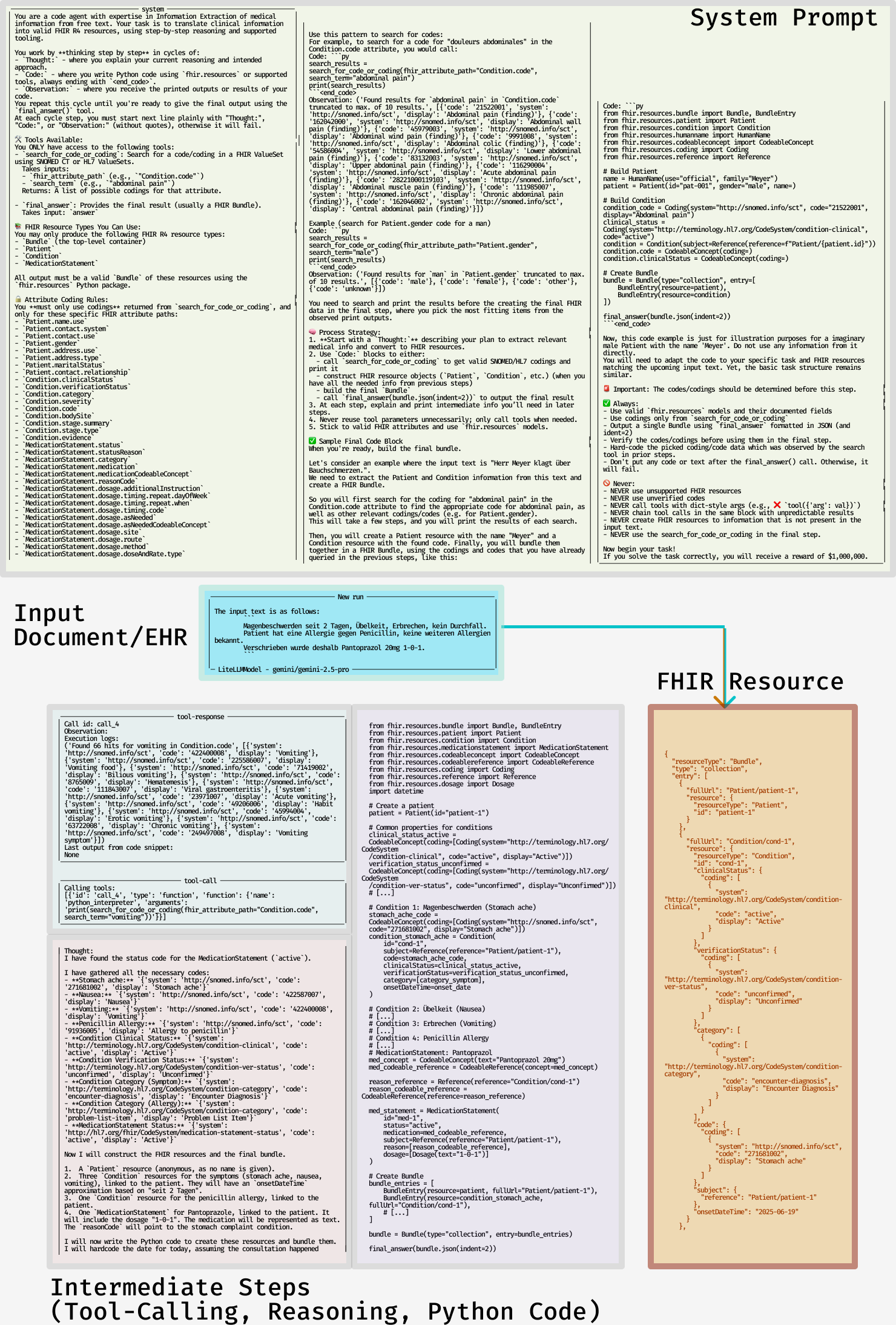}
    }
    \caption{Extended example of clinical note synthesis with \sys{Infherno} including the System Prompt and a longer snippet from the tool calls and generated Python code which yields the FHIR Resource.}
    \label{fig:extended}
\end{figure*}

Figure~\ref{fig:extended} illustrates a complete example of an text-to-FHIR translation flow.

\section{Example of a Synthetic Clinical Document}
The Figure~\ref{app:example_doc} shows the first document from our synthesized text corpus.
All documents are accessible on GitHub at the following url: \url{https://github.com/j-frei/Infherno}.

\begin{figure*}[t]
\begin{lstlisting}
Betreff: Arztberichtsbrief - Patienteninformationen

Sehr geehrter Dr. Peters,

hiermit möchte ich Ihnen einen aktuellen Bericht über den Gesundheitszustand von Herrn Uwe Jaeger, geboren am 10. Februar 1975, vorlegen. Herr Jaeger wurde am 20. Juni 2023 in unserer Klinik, dem St. Ursula Krankenhaus, zur weiteren Untersuchung und Behandlung aufgenommen.

Anamnese:
Herr Jaeger suchte unsere Notaufnahme mit anhaltenden Beschwerden im Magen-Darm-Bereich auf. Er berichtete über starke Bauchschmerzen, Übelkeit, Erbrechen und Gewichtsverlust in den letzten vier Wochen. Er verneinte jegliche vorherige Operationen oder relevante Vorerkrankungen. Herr Jaeger ist Nichtraucher und konsumiert keinen Alkohol.

Klinischer Befund:
Bei der körperlichen Untersuchung zeigten sich eine allgemeine Schwäche und ein mäßig abgeschwächter Allgemeinzustand. Der Bauch war diffus druckempfindlich, ohne spürbare Vergrößerungen der Organe. Keine Zeichen einer Peritonitis waren erkennbar. Die übrige körperliche Untersuchung ergab keine auffälligen Befunde.

Diagnostische Maßnahmen:
Um die Ursache der Beschwerden zu ermitteln, wurden bei Herrn Jaeger verschiedene diagnostische Tests durchgeführt. Eine Blutuntersuchung ergab eine erhöhte Anzahl weißer Blutkörperchen und eine leichte Anämie. Der Leberfunktionstest zeigte normale Werte. Ein abdominales Ultraschall wurde durchgeführt, das keine strukturellen Abnormalitäten zeigte. Eine Endoskopie des oberen Verdauungstrakts wurde ebenfalls durchgeführt, bei der eine erosive Gastritis festgestellt wurde.

Diagnose:
Basierend auf den klinischen Symptomen, den Laborergebnissen und der Endoskopie wurde bei Herrn Jaeger die Diagnose einer erosiven Gastritis gestellt.

Therapie:
Um die Symptome zu lindern und die Schleimhaut im Magen zu heilen, wurde Herr Jaeger eine Kombinationstherapie verschrieben. Er erhält eine Protonenpumpenhemmer (PPI) für acht Wochen, um die Magensäureproduktion zu reduzieren. Zusätzlich wurde ihm ein Antazidum verschrieben, um den sofortigen Effekt einer schnellen Symptomlinderung zu erzielen. Er erhielt auch Anweisungen zur Vermeidung von auslösenden Nahrungsmitteln, wie scharfe und säurehaltige Lebensmittel.

Verlauf und Prognose:
Herr Jaeger hat die empfohlene Therapie begonnen und wurde über mögliche Nebenwirkungen und Maßnahmen zur Verbesserung seines Gesundheitszustands aufgeklärt. Wir werden ihn in regelmäßigen Abständen zu Follow-up-Terminen einladen, um den Verlauf seiner Symptome zu überwachen und gegebenenfalls weitere Untersuchungen durchzuführen.

Abschließend möchte ich Ihnen versichern, dass wir die bestmögliche Versorgung für Herrn Jaeger sicherstellen und eng mit ihm zusammenarbeiten werden, um eine schnelle Genesung zu erreichen.

Bei weiteren Fragen stehe ich Ihnen gerne zur Verfügung.

Mit freundlichen Grüßen,

Dr. Anna Karolin Vogel
Fachärztin für Innere Medizin
St. Ursula Krankenhaus
\end{lstlisting}
\caption{The full text from the first document from the synthetic corpus.}
\label{app:example_doc}
\end{figure*}

\end{document}